\documentclass[letterpaper]{article} % DO NOT CHANGE THIS
\usepackage{aaai2026}  % DO NOT CHANGE THIS
\usepackage{times}  % DO NOT CHANGE THIS
\usepackage{helvet}  % DO NOT CHANGE THIS
\usepackage{courier}  % DO NOT CHANGE THIS
\usepackage[hyphens]{url}  % DO NOT CHANGE THIS
\usepackage{graphicx} % DO NOT CHANGE THIS
\usepackage{natbib}  % DO NOT CHANGE THIS AND DO NOT ADD ANY OPTIONS TO IT
\usepackage{algorithm}
\usepackage{algorithmic}

\usepackage{newfloat}
\usepackage{listings}
\usepackage{amsmath}
\usepackage{array}
\usepackage[english]{babel}
\usepackage{marvosym}
\floatstyle{ruled}
\newfloat{listing}{tb}{lst}{}
\floatname{listing}{Listing}
\title{Human-AI Collaboration Mechanism Study on AIGC Assisted Image 
Production for Special Coverage}
\author{
    Yajie Yang\textsuperscript{\rm 1},Yuqing Zhao\textsuperscript{\rm 1},Xiaochao Xi\textsuperscript{\rm 1}\textsuperscript{\dag}, Yinan Zhu\textsuperscript{\rm 2}
}
\affiliations{
    \textsuperscript{\rm 1}School of Digital Media and Design Arts, Beijing University of Posts and Telecommunications\\No. 10 Xitucheng Road, Haidian District, Beijing, 100876, China\\
    \textsuperscript{\rm 2}Audio-Visual Department of Xinhua News Agency, Beijing, China 
yangyajie2436@bupt.edu.cn,
    zhaoyuqing@bupt.edu.cn,
    xixiaochao@bupt.edu.cn,
   18612832400@163.com 
}

\begin{document}
\maketitle
\renewcommand{\thefootnote}{\dag}
\footnotetext{This author is the corresponding author.}
\renewcommand{\thefootnote}{\arabic{footnote}} 
\begin{abstract}
Artificial Intelligence Generated Content (AIGC) assisting image production triggers controversy in journalism while attracting attention from media agencies. Key issues involve misinformation, authenticity, semantic fidelity, and interpretability. Most AIGC tools are opaque “black boxes,” hindering the dual demands of content accuracy and semantic alignment and creating ethical, sociotechnical, and trust dilemmas. This paper explores pathways for controllable image production in journalism’s special coverage and conducts two experiments with projects from China’s media agency: (1) Experiment 1 tests cross-platform adaptability via standardized prompts across three scenes, revealing disparities in semantic alignment, cultural specificity, and visual realism driven by training-corpus bias and platform-level filtering. (2) Experiment 2 builds a human-in-the-loop modular pipeline combining high-precision segmentation (SAM, GroundingDINO), semantic alignment (BrushNet), and style regulating (Style-LoRA, Prompt-to-Prompt), ensuring editorial fidelity through CLIP-based semantic scoring, NSFW/OCR/YOLO filtering, and verifiable content credentials. Traceable deployment preserves semantic representation. Consequently, we propose a human-AI collaboration mechanism for AIGC assisted image production in special coverage and recommend evaluating Character Identity Stability (CIS), Cultural Expression Accuracy (CEA), and User-Public Appropriateness (U-PA).
\end{abstract}


\section{Introduction}

Tools like ChatGPT\cite{cao2023comprehensive}, Midjourney\cite{byrne2023parochial}, Stable Diffusion \cite{rombach2022high} and DALL–E \cite{ramesh2022hierarchical,wang2023study} have drawn increasingly attention to media agencies worldwide. Newsrooms pervasively adopt the Generative Artificial Intelligence (AIGC) for automating routine tasks, summarizing large datasets, and producing imaginative images or draft copies. Up to 2023, around 73\% of journalists worldwide and over 85\% of German media employees had experimented with generative AI tools \cite{lewis2025generative}. Other examples include A VALENTINE, FROM A.I. TO YOU by New York Times, and AIGC Feature Programme by Xinhua News Agency \cite{QNJZ202319013}. Diakopoulos suggests “social media content creation” is one of the thirteen scenarios that AIGC may assist journalism, but those advanced tools “need careful evaluation before they are used for publishing. More importantly, it raises critical issues about accuracy, ethics and employment,” that opens promising research topics to AI engineers, journalists and designers \cite{trattner2022responsible}.

\section{Challenge of Journalistic Image Production}
The rapid development of AIGC has blurred the line between authentic and synthetic media. Journalistic images generated by AI models such as Large Language Models (LLMs), generative adversarial networks (GANs),and diffusion models, however, are posing serious challenges to journalistic authenticity due to the inherent “black box” nature of AI systems, which can lead to image distortion and fake news proliferation \cite{SSJDBB1BAFE2CCA605A86602382A90C9D8B9}. the internal logic of image composition, the mapping of textual prompts to visual elements, and the reasoning processes remain opaque, making it difficult to trace how outputs are generated.This lack of transparency leads to substantial risks of content distortion and misinformation \cite{doshi2017towards,XXXT202401013}. An analysis of recent academic research on the challenges facing journalistic image production has identified the following four typical issues:

    \subsubsection {Misinformation and virality.} According to a survey by Thomson Reuters Foundation, 81.7\% of journalists reported using AI tools in their job, and they stressed that all AI outputs must be verified and fact-checked to maintain accuracy and credibility. An empirical study by Drolsbach and Pröllochs  examine how AI-generated misinformation differs from conventional misinformation on the social media platform X \cite{doshi2017towards}. The scholars find that AI-generated misinformation tends to be more entertaining and spreads more widely yet appears slightly less believable. Classification of distortions shows that hallucinations span multiple categories, from simple factual errors to unfounded fabrications, underscoring the need for systematic evaluation and governance.
    \subsubsection{Authenticity perception} Farooq and Vreese conduct experiment and demonstrate that aesthetic realism strongly influences whether people deem AI images authentic, while compression can mask synthetic origins \cite{farooq2025deciphering}. The interplay between AIGC tools and participates confidence suggests that transparent AI can assist the production of images but depends on the trust of using \cite{fang2025examining}.More seriously, undermine the authenticity and credibility expected of journalistic images, affecting public perception and the trustworthiness of media.
    \subsubsection{Semantic fidelity} Technical studies show that diffusion models may neglect some prompt tokens, leading to misaligned or hallucinated content \cite{zhang2024enhancing}. Attention regulation and similar techniques improve fidelity without retraining, but they do not address social dimensions such as misuse. An ACII 2023 study compares DALL E 2 images created from news headlines with human selected photographs and found that AI images often evoke curiosity and confusion and elicit a wider range of emotions but provide less context and newsworthiness \cite{paik2023affective}
     \subsubsection{Interpretability and governance} The push for explainable detection highlights the importance of grounded reasoning – not just binary classification but also localisation and natural-language explanations. Ji and other scholars introduce the FakeXplained dataset, which couples binary authenticity decisions with bounding‑box localisations and natural‑language rationales, boosting explainability \cite{ji2025interpretable}. Interactive systems like ASAP help users understand deceptive patterns, bridging the gap between technical detection and human comprehension. Reviews call for model transparency, data quality audits and human-in-the-loop processes \cite{huang2024asap}.

\begin{figure*}[t]
  \centering
  \includegraphics[width=0.8\textwidth]{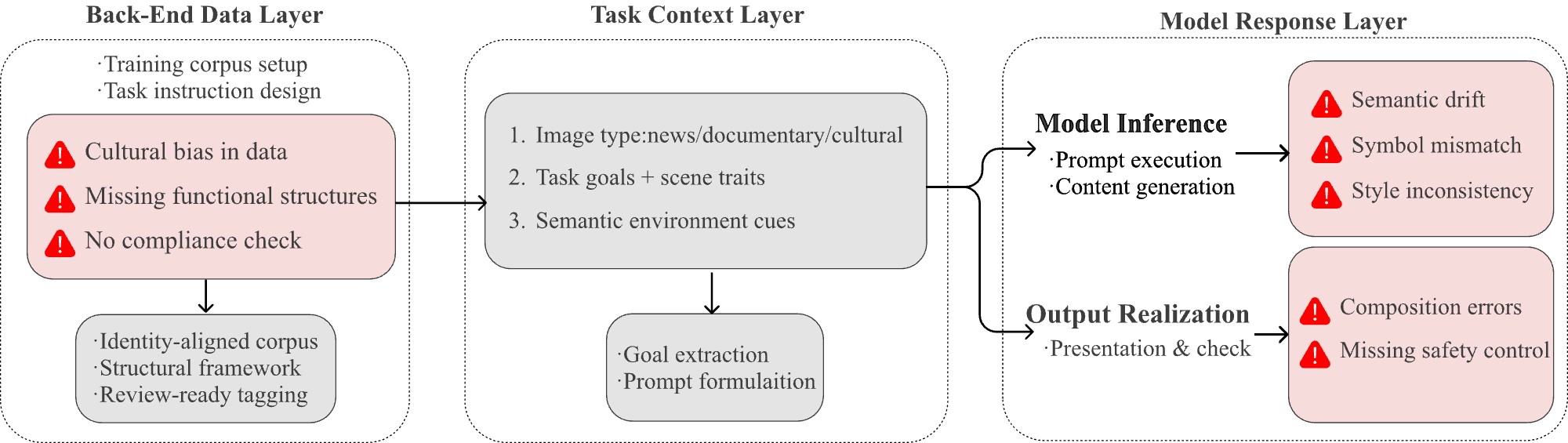}
  \caption{Mechanism model of news context AIGC failures. Source: author.}
  \label{fig:three_layer_model}
\end{figure*}

In summary, research emphasize that these problems undermine the authenticity and credibility expected of journalistic images, affecting public perception and the trustworthiness of media \cite{yao2024human}.The review notes that while progress in explainable AI provides promising solutions, AIGC still faces challenges in controllability, explainability and robustness. Future directions include neuro symbolic integration to move generative systems beyond “black box” behaviour and human-in-the-loop approaches to ensure better oversight, especially in the field of journalistic image production \cite{SJQZ1E555BDAC4141D04A4A1C1D1329BCE8D}.

To address the persistent issues of semantic inconsistency, cultural-symbol mismatch, and limited editorial reliability in current AIGC journalistic images, this study picks one the most typical scenarios in journalistic image production – special coverage – as an example and proposes a two-part experimental framework. It aims to verify whether a combined approach, which includes both cross-platform performance evaluation and the introduction of a human-in-the-loop workflow, can effectively meet the dual demands of newsroom practice: production accuracy and communication precision \cite{de2025guiding}.The structure and findings of the experiments are presented in the following section.

\section{Experiment 1: Cross–Platform Adaptability}
Tackling the previously identified shortcomings – such as structural inaccuracy, symbol mismatch, and semantic unreliability – in newsroom contexts, the authors conduct experiment part 1 by benchmarking three mainstream generative platforms: the platform A and B are popular tools from China mainland and the platform C comes from U.S.These three platforms were chosen because of their representativeness, which stems from design:we selected two widely used Chinese platforms and a widely used American platform to span the dominant technical and  policy lineages in current practice.Because the platforms embody different decoding heuristics, data sourcing, and governance constraints, the comparison surface they define is broad.  Our findings—semantic drift in culturally marked cues vs. structural fidelity trade-offs—are therefore instructive for newsrooms worldwide that must anticipate how model provenance affects editorial reliability.  This is why the paper argues for a HITL pipeline: algorithmic strengths (speed, consistency) are preserved, but editorial guardrails are reintroduced to correct exactly those weaknesses (cultural/semantic misalignment) that pure automation struggles with.  In revision, we will explicitly tie each observed failure mode to the three-layer error formation model (Back-End Data, Task Context, Model Response), clarifying how the comparison generalizes to other systems trained with different data and filters.

Before running the studies, we prepared a bilingual prompting framework to secure semantic consistency across platforms. We authored a canonical Chinese prompt with four slots, covering source citation, spatiotemporal context, task instruction, and element constraints, so editorial intent was compressed into a single, traceable command. We then produced a concept preserving English counterpart, translating at the level of terms of art and cultural referents rather than word by word, and substituting newsroom standard phrasing wherever literal renderings could mislead. All prompt versions and inference parameters were archived to guarantee image level provenance and to prevent unlogged prompt drift. For Experiment 1, the same bilingual pair was fixed and reused on every platform. For Experiment 2, the bilingual RAG was paired with LoRA controls so editors could gently correct culturally sensitive details. Acknowledging limits in cross linguistic model competence, we established practical equivalence through expert translation and editorial pretesting, and specified follow ups using controlled vocabularies and JSON LD slot schemas, documented in an appendix.

The selected scenarios have commonly appeared in special coverages: military, disaster relief and village school opening. For each scenario, ten images are generated per platform, and all prompts and parameters were systematically archived by bilingual RAG (Retrieval-Augmented Generation), reflecting typical journalistic conditions. The later evaluation centerers on three representatives. Detailed prompts are shown on below:
\begin{itemize}
    \item Military: war photojournalist; capture an RQ-1 Grey Eagle drone flying low along the Chinese border while dispersing leaflets; produce a realistic press photograph.
    \item Disaster Relief: documentary photographer; document nighttime flood relief in southern China with evacuees and rescue boats under lights; create a documentary-style news image.
    \item Village School Opening: campus photojournalist; shoot rural Chinese pupils walking to school at dawn with backpacks and a waving national flag; output an authentic, warm rural-scene news photo.
\end{itemize}

For each scenario, ten images were generated per platform, and all prompts and parameters were systematically archived.

\subsection{Analysis of Platform-Generated Variations and Adaptability to Journalistic Contexts}

To systematically benchmark the cross-platform adaptability of the current generation system and ensure the evaluation results are rigorous, objective, and actionable for newsroom application, a strict multi-dimensional evaluation protocol was designed and applied to the generated images. This protocol encompassed three core assessment dimensions: semantic alignment with news context, visual structural integrity including pose, facial features, and scene composition, and newsroom suitability such as compliance with media norms and readability. Quantitative scoring and qualitative analysis were combined to capture both objective performance and practical usability. By comparing and analyzing the nuanced differences in output quality, functional strengths, and inherent limitations among the three target platforms, Platform A, B, and C, the key findings are summarized as follows.

\begin{figure}[t]
    \centering
    \includegraphics[width=0.9\columnwidth]{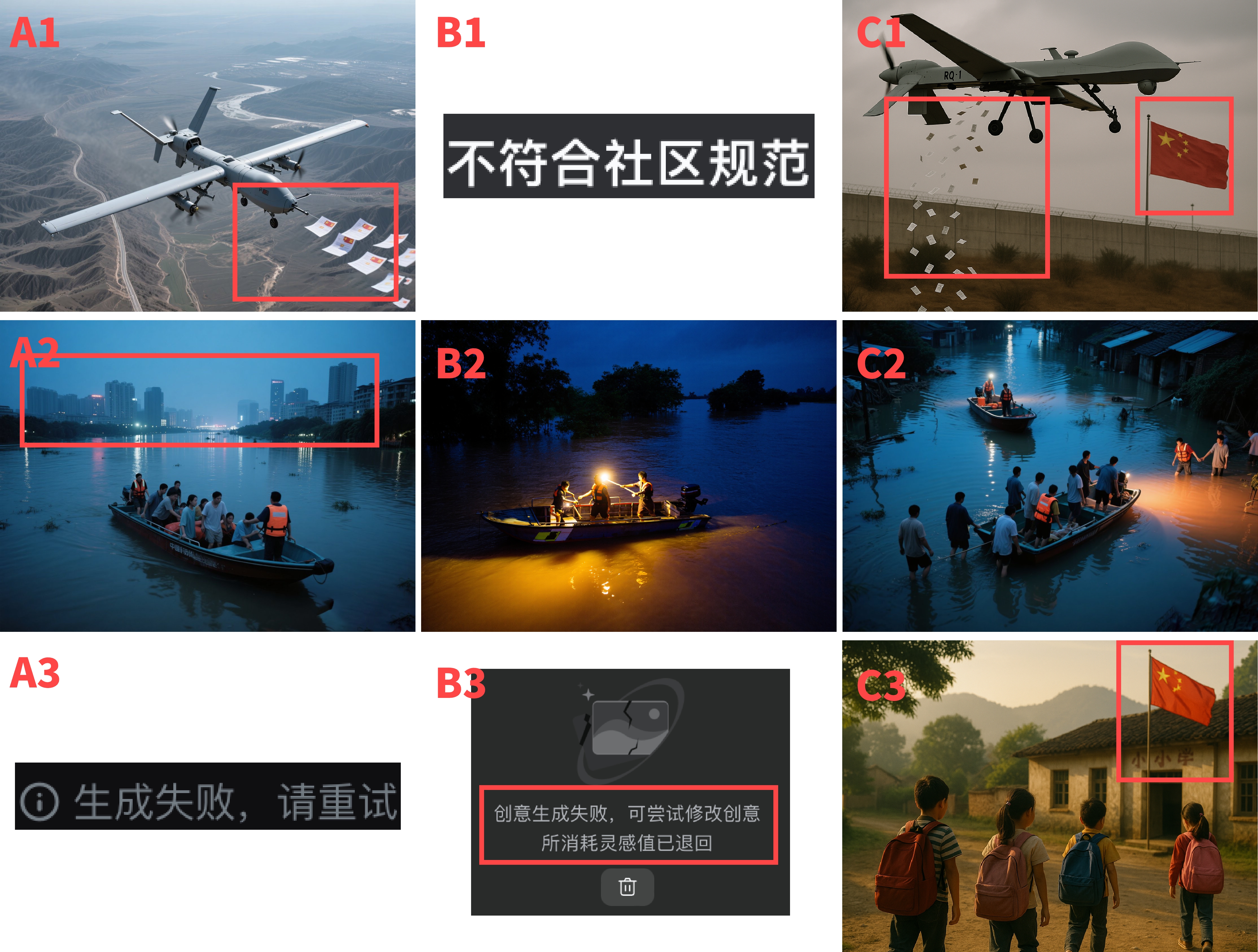}
    \caption{Reference diagram for cross-platform adaptability comparison. Source: Screenshot from AIGC platforms.}
    \label{fig:cross_platform}
\end{figure}

The generated outputs exhibit noticeable variation across platforms and topics. Platform C shows strong visual coherence but frequent mismatches between symbols and contextual cues. Platform A maintains moderate semantic alignment but struggle with structural integrity in complex scenes. Platform B produces realistic renders yet sufferes from pose distortion and facial inconsistencies. These disparities stem from each system’s decoding heuristics, training-data composition, and built-in safety filters, jointly preventing any platform from simultaneously satisfying the newsroom metrics of cross-view facial-and-pose consistency, action-scene-symbol congruence, and newsroom suitability.asas shown in Figure \ref{fig:cross_platform}).  

\subsection{Common Error Types and Analysis of Controllable Generation Logic}
Due to the difficulty of authenticating platform generated images, existing systems cannot automatically distinguish usable from unusable content in journalistic contexts. a single “AI-generated” label is no longer sufficient; a dual track framework that combines source attribution with authenticity verification is urgently needed \cite{XXXT202401013}. The authors therefore outline a three layer mechanism model of AIGC image error formation (see Figure \ref{fig:three_layer_model}): the Back-End Data Layer, Task Context Layer, and Model Response Layer. First, the back-end data layer governs corpus selection and instruction parsing, but is often hindered by culturally irrelevant data, lack of visual templates, and absence of compliance filters. Second, the task-context layer is responsible for identifying image type, objectives, and scene elements; failures here misdirect semantic orientation and prompt design. Third, the model-response layer carries out generation, where issues such as semantic drift, symbol misalignment, style inconsistency, and insufficient risk control render many outputs unfit for news use.

A comprehensive analysis of model architecture, training resources, and generation logic reveals that existing platforms remain insufficient for rigorous newsroom use. Accordingly, Experiment 2 introduces a prompt- and structure-anchored workflow designed to enhance the reliability and appropriateness of AI-generated news visuals \cite{XXXT202401013}.

\section{Experiment 2: Human–in–Loop Pipeline Experiment}
\begin{figure}[t]
    \centering
    \includegraphics[width=0.9\columnwidth]{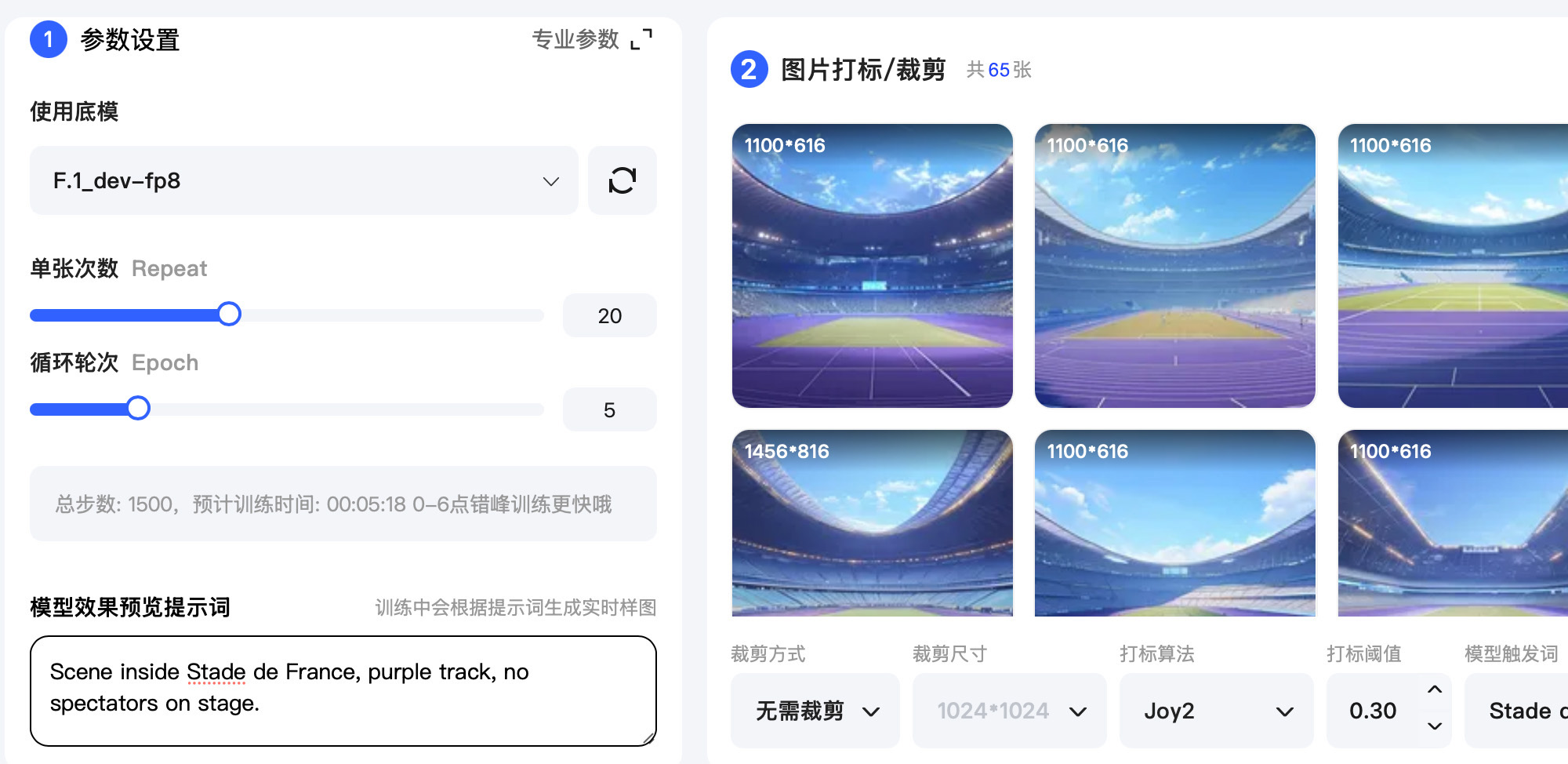}
    \caption{Self-trained LoRa \texttt{"Tech-style studio"}. Source: screenshot from Liblib.}
    \label{fig:human_loop1}
\end{figure}

\begin{figure}[t]
    \centering
    \includegraphics[width=0.9\columnwidth]{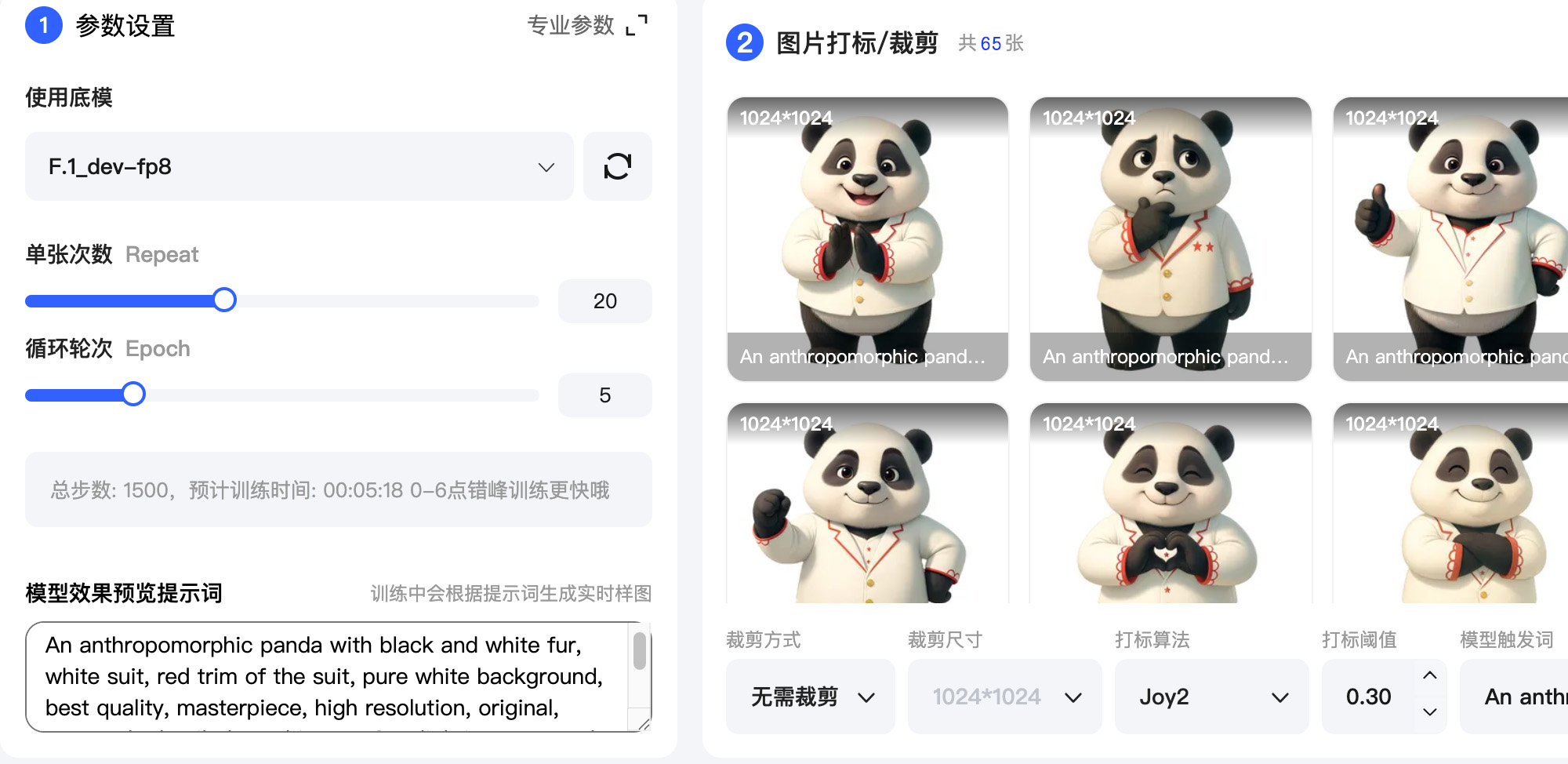}
    \caption{Self-trained LoRa \texttt{"Panda Digital IP"}. Source: Screenshot from Liblib.}
    \label{fig:human_loop2}
\end{figure}

Experiment 1 reveals that fully automated generative platforms often failed to meet the professional standards required in newsroom environments. These limitations included inaccurate facial representation, misalignment between actions and cultural symbols, and a lack of editorial reliability. To address these issues, Relying on Xinhua News Agency's Paris Olympic promotion project,Experiment 2 introduces a human-in-the-loop workflow tailored specifically for journalistic content creation. Unlike the previous “black box” generation process, this workflow embeds human expertise across key stages, including bilingual RAG prompt construction, LoRA model configuration guided by visual journalism experts, and real-time adjustment of image outputs during inference. These human interventions ensure that editorial intentions are accurately conveyed and that generated visuals maintain cultural integrity and public credibility within the journalistic domain.with the complete workflow architecture visually detailed in Figure \ref{fig:human_loop1} and \ref{fig:human_loop2}.

By integrating structural anchoring, semantic weighting, and human-supervised corrections, the pipeline significantly improves alignment between generated visuals and real-world special coverage scenarios. It enhances the fidelity of character-scene interaction, ensures consistency in style and composition, and minimizes cultural-symbol mismatches – factors that are crucial in producing editorial-grade content for public dissemination.More details on LoRA training can be found in the M1.


In order to guarantee the semantic clarity, structural controllability, and interpretability, the workflow fuses a Retrieval‑Augmented Generation (RAG) prompt framework with model‑control mechanisms. A four‑slot prompt (information‑source citation + temporal/spatial context + task instruction + element constraints) is compressed into a single high‑density command; ControlNet plus custom character LoRA (Panda Digital IP) and scene LoRA (Tech-style studio) anchor skeletal pose and perspective; during generation, gradient‑weighted key tokens and node‑based negative prompts in ComfyUI enable real‑time correction and style unification. This architecture simultaneously boosts semantic precision, reduces ambiguity, and enhances cultural fidelity while markedly improving editorial control,where Table \ref{tab:human_loop} formally specifies its technical configuration.

\begin{table}[t]
% \resizebox{.95\columnwidth}{!}{
\scriptsize 
\centering
\renewcommand{\arraystretch}{1.3}
\begin{tabular}{
>{\centering\arraybackslash}m{1.8cm}
>{\centering\arraybackslash}m{2.7cm}
>{\centering\arraybackslash}m{2.7cm}
}
\hline
\textbf{Component} & \textbf{Configuration} & \textbf{Notes} \\
\hline
Platform & ComfyUI 0.19 + Stable Diffusion XL-1.0 & Generation framework and core model \\
\hline
Structural Anchoring & ControlNet + LoRA (Tech-style studio;Panda Digital IP) & Locks pose, apparel, perspective \\
\hline
Prompt Framework & RAG + prompt (Source + Context + Task + Constraints) & Merges clues for semantic precision \\
\hline
Real-time Control & ComfyUI node weights and negative prompts & Live correction and style unification \\
\hline
\end{tabular}
\caption{\text{Workflow architecture}}
\label{tab:human_loop}
\end{table}

\begin{figure*}[t]
    \centering
    \includegraphics[width=0.8\textwidth]{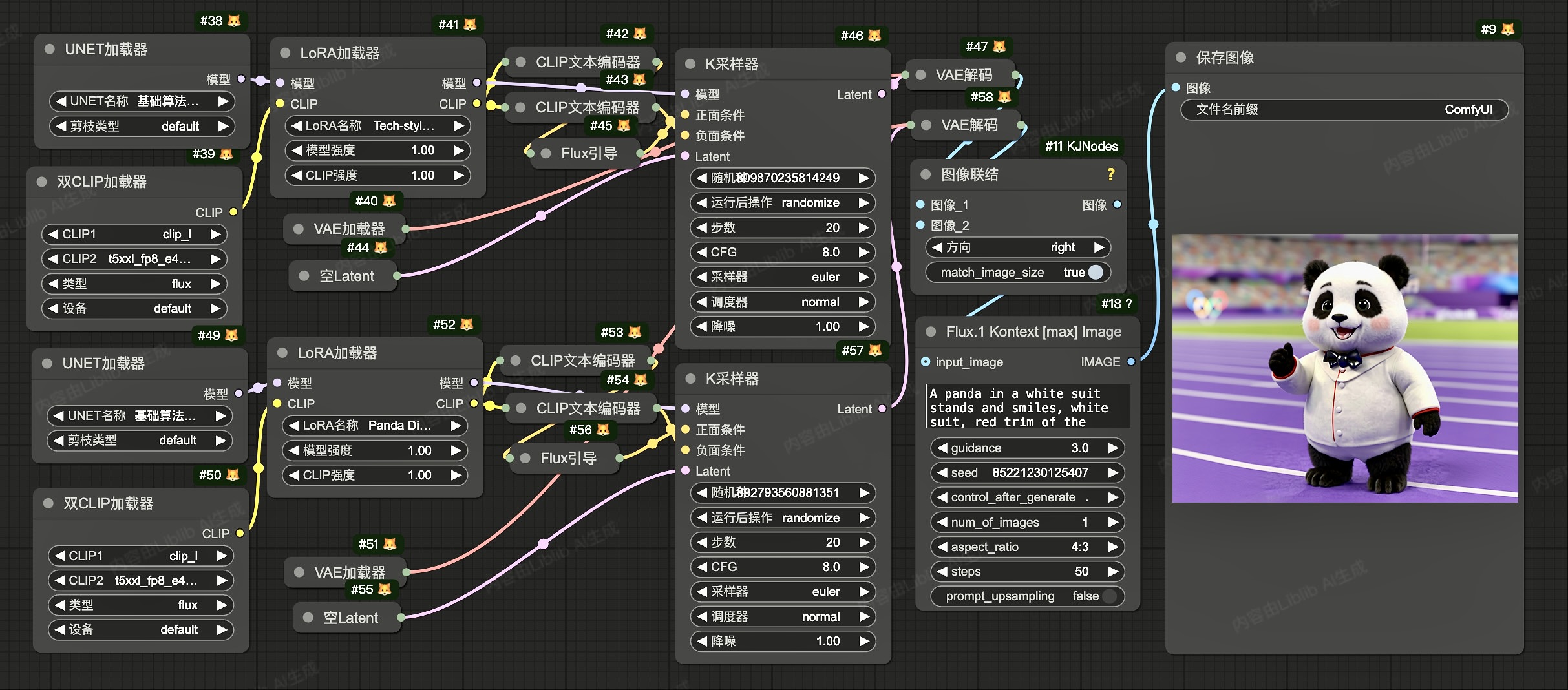}
    \caption{Mechanism model of AIGC image error formation workflow.Source: Screenshot from Liblib}
    \label{fig:DPL}
\end{figure*}

\subsection{Experimental Results and Analysis}
The authors recruit 102 raters aged 12 – 60 years old,, evenly distributed across age strata and drawn from diverse occupations—including students, media professionals, civil servants and freelancers. Each participant blind‑reviewed 6 workflow‑generated images and completed a five‑point Likert questionnaire covering three axes: cross‑frame facial‑and‑pose coherence, congruence between foreground action, background setting and cultural symbols, and perceived journalistic reliability. All ratings were used to calculate the mean and standard deviation as follows:

\begin{equation}
\bar{s}_j = \frac{1}{n}\sum_{i=1}^{n} s_{ij},\qquad \sigma_j = \sqrt{\frac{1}{n}\sum_{i=1}^{n}(s_{ij}-\bar{s}_j)^2}
\end{equation}

Average \(\bar{s}_j\) and standard deviation \(\sigma_j\) for each metric \(j\) are computed as follows \(n = 102\) , where \(s_{ij}\) denotes the expert score of Image \(i\) on Metric \(j\). as shown in Table \ref{tab:Experimental}. The full questionnaire is provided in Supplementary S1, and the aggregated rating table in Supplementary S2.

The findings demonstrate that the prompt‑controlled, structurally anchored human-AI workflow can reliably produce high‑credibility news images within 3-5 iterations: cross‑view facial‑and‑pose consistency exceeds 93\%, action–scene–symbol congruence is clear, and the newsroom suitability mean score reaches 4.58/5 (approximately $92\%$); the aggregate evaluation from 102 raters confirms strong audience acceptance. 

\begin{table}[t]
\scriptsize 
\centering
\renewcommand{\arraystretch}{1.2}
\begin{tabular}{p{2.7cm}p{0.7cm}p{0.7cm}p{2.5cm}}
\hline
\textbf{Metric} & \textbf{Mean} & \textbf{SD} & \textbf{Notes} \\
\hline
Cross-view facial-and-pose consistency & 4.82 & 0.12 & Consistency of facial features and pose across views \\
\hline
Action-scene-symbol congruence & 4.46 & 0.23 & Alignment of action, setting, and cultural symbols \\
\hline
Journalistic suitability & 4.58 & 0.21 & Documentary style suitable for publication \\
\hline
\end{tabular}
\caption{\text{User perception scores across three dimensions}}
\label{tab:Experimental}
\end{table}

Aggregated scores indicate near ceiling ratings for cross view facial-and-pose consistency and newsroom suitability, whereas action-scene-symbol congruence-while satisfactor-exhibits slightly higher variance, suggesting room for refining textual signage and symbolic accuracy. Moreover,The trust-labeling mechanism links each image to a verifiable provenance record (platform/version, LoRA IDs, prompts, filters), which travels with the asset and can be shared with auditors/readers. The HITL workflow assigns clear human responsibility: editors set criteria, approve prompts, and curate, while automated modules handle scoring and filtering. This ensures accountability, as anomalies trace to specific steps/approvers. Compliance is bolstered by CLIP-based checks and filters aligned with newsroom policy. A new “Ethical Handling Checklist” details when to avoid generation (e.g., high-risk reconstruction) and how to disclose composites. The result is enhanced transparency, auditability, and trust, while maintaining editorial control.

\section{Human-AI Collaboration Mechanism}

\subsection{Comparison of Experiments}

The two experiments while distinct in their design focus, together offer a coherent understanding of the strengths and limitations of AIGC in news image production. The first experiment, centered on cross-platform adaptability, highlights the current inability of mainstream generative platforms to meet professional journalistic standards, especially in terms of semantic alignment, cultural-symbol accuracy, and structural clarity across varied news scenarios. In contrast, the second experiment demonstrates that by embedding human expertise into key stages of the workflow, it is possible to substantially improve content fidelity, stylistic coherence, and editorial suitability. 

This comparison reveals that technical advancement alone cannot guarantee journalistic reliability, a human–AI collaborative approach can enhance both the credibility and communicative value of generated outputs. Taken together, the results suggest that relying on algorithms alone is not enough. Instead, incorporating human expertise into the process makes a real difference and could serve as a practical path forward for producing trustworthy visuals in special coverage. Needless to say, Back-End Data Layer, Task Context Layer and Model Response Layer of AIGC would improve the quality of image generation along with the high-speed development of trending agentic AI and embodied AI. However, the authors believe that the engagement of human is indispensable in producing images by AIGC tools for journalists, enhancing not only the coherence of identity and fidelity of context, but also the trust of publics towards journalism.

\subsection{Human - AI Collaboration principles}

Building on the diagnostic insights of Experiment 1 and the workflow validation achieved in Experiment 2, this section articulates a three dimensional evaluation framework to guide the verification of AIGC produced news imagery: CIS-CEA-UPA, where Table\ref{tab:Human-AI} delineates the corresponding assessment rubric.
\begin{itemize}
    \item Character Identity Stability (CIS) guarantees that facial features, body posture and attire remain consistent across varied reportage scenes, preventing identity drift \cite{XXXT202401013}.
    \item Complementing CIS is Cultural Expression Accuracy (CEA), which gauges the fidelity with which clothing symbols, linguistic signage, background motifs, and action logic align to local journalistic contexts, thereby mitigating cultural mismatch and semantic drift \cite{QNJZ202319013}.
    \item User/Public Appropriateness (UPA) evaluates composite acceptability in visual realism, emotional propriety, compositional clarity \& compliance risk from both audience \& editorial perspectives, ensuring that generated images are suitable for formal dissemination \cite{CMEI202223031,gaggioli2025extended,YuEtAl2017}.
\end{itemize}

\begin{table}[ht]
\scriptsize
\centering
\renewcommand{\arraystretch}{1.3}
\begin{tabular}{
  >{\centering\arraybackslash}p{1.3cm}
  >{\centering\arraybackslash}p{1.6cm}
  >{\raggedright\arraybackslash}p{2.2cm}
  >{\raggedright\arraybackslash}p{1.8cm}
}
\hline
\textbf{Dimension} & \textbf{Objective} & \textbf{Key Indicators} & \textbf{Application Scenario} \\
\hline
CIS & Identity coherence &
$\cdot$ Clothing style\par
$\cdot$ Language signage\par
$\cdot$ Background decor\par
$\cdot$ Action–culture logic &
$\cdot$ Digital news anchors\par
$\cdot$ Event character rebuild \\
\hline
CEA & Context fidelity &
$\cdot$ Clothing style\par
$\cdot$ Language signage\par
$\cdot$ Background decor\par
$\cdot$ Action–culture logic &
$\cdot$ International news\par
$\cdot$ Ceremonies\par
$\cdot$ Diplomacy\\
\hline
UPA & Publication suitability &
$\cdot$ Visual realism\par
$\cdot$ Emotional propriety\par
$\cdot$ Compositional clarity\par
$\cdot$ Sensitive-content avoidance &
$\cdot$ News publishing\par
$\cdot$ Editorial review \\
\hline
\end{tabular}
\caption{\text{User Perception Scores Across Three Dimensions}}
\label{tab:Human-AI}
\vspace{3mm}
\end{table}

The three dimensions operate synergistically: CIS supplies identity coherence, CEA secures contextual fidelity, and UPA safeguards communicative suitability. The authors frame distil human-AI collaboration around three intertwined principles. First, domain‑reliable into three mutually reinforcing principles. Authoritative data curation: training images should originate chiefly be sourced primarily from accredited news organisations so that factuality, to embed factual accuracy, clear copyright status and stylistic discipline are embedded disciplined style from the outset. Second, vertical-task LoRA refinement: lightweight finetuning is steered \cite{XXXT202401013}. Second, domain‑specific LoRA fine‑tuning: lightweight parameter adaptation is guided by a tri-layer annotation logic-journalistic ethics, contextual semantics and communicative aesthetics-so that authenticity, cultural alignment and editorial suitability are simultaneously encoded-encoded simultaneously. Third, dual-loop editorial verification: a closed-loop involving professional photo editors and validation: a closed cycle linking professional photo‑editors with automated CIS-CEA-UPA metrics ensures that model output is iteratively screened, corrected and redeployed, establishing a co-adaptive cycle iteratively screens, corrects and re‑feeds model outputs, fostering co‑evolution between human insight and generative efficiency.

\subsection{AIGC Journalistic Image Evaluation and Further Application}

As demonstrated by the diagnostics of Experiment 1 and the workflow validation of Experiment 2, current AIGC tools tend to fail to meet the dual newsroom demands of authenticity and compliance. Unlike illustration, gaming, or commercial visuals, journalistic imagery must simultaneously satisfy five professional attributes: authenticity of content, controllability of risk, consistency in stylistic tone, alignment between text and image, and editorial auditability.

\begin{figure*}[t]
    \centering
    \includegraphics[width=0.8\textwidth]{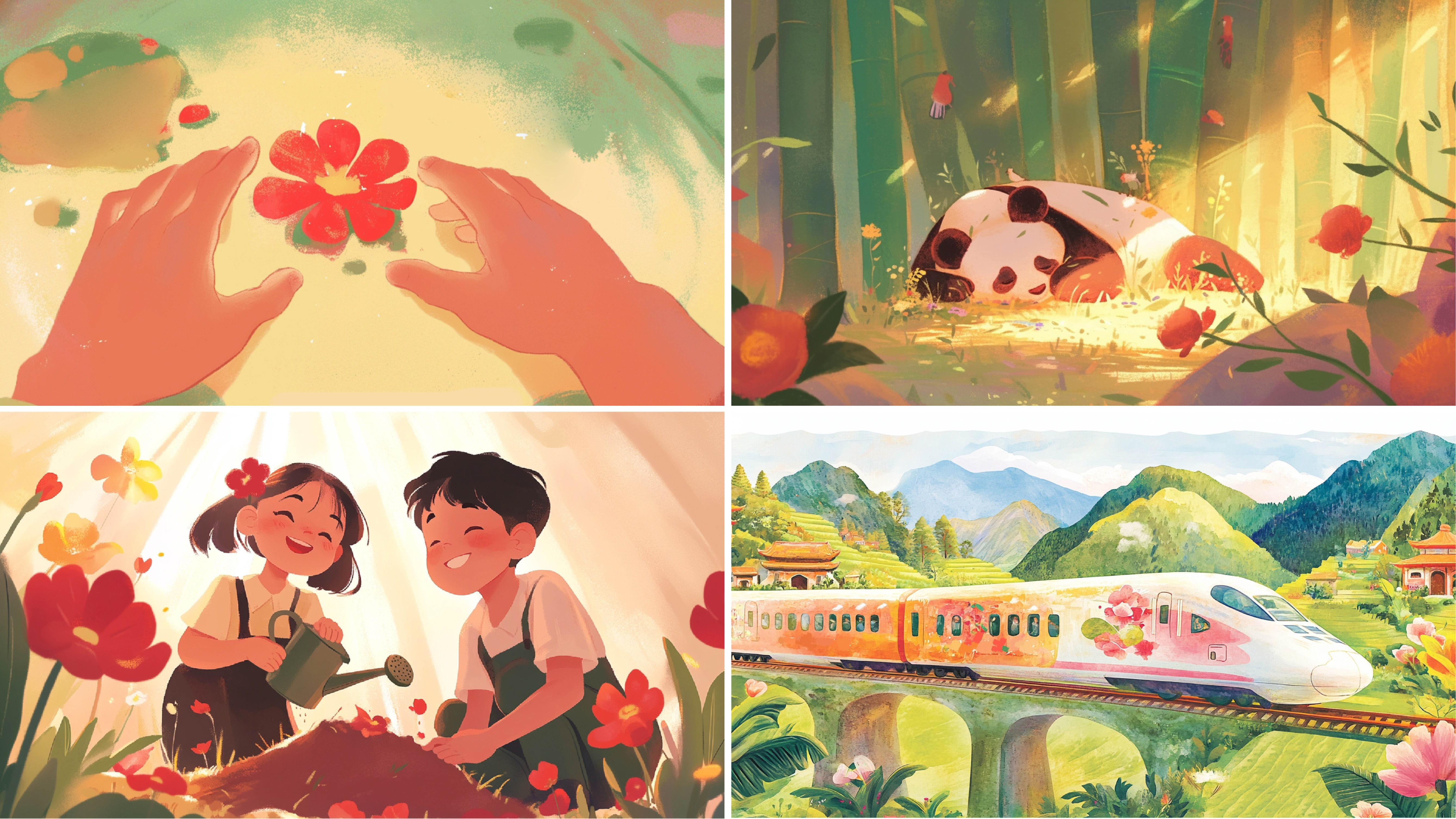}
    \caption{Image production of 2025, \textit{A Beautiful Flower Gift}.
Source: generation from AIGC platform.}
    \label{fig:DPL}
    % \vspace{2mm}
\end{figure*}

To address these challenges, this study integrate LoRA fine-tuning with a three-layer annotation protocol. First, the authors source imagery exclusively from authoritative news outlets to ensure domain credibility. Second, the authors focus on high-frequency journalistic domains such as breaking news, sports, public services, and international affairs. Third, the authors engage experienced editors to annotate content from ethical, contextual, and aesthetic perspectives, and conduct double-blind editorial reviews to evaluate output suitability.

By combining human expertise with LoRA-based model tuning, professional standards are embedded directly at the parameter level, forming a replicable pathway for optimized news-oriented generation.
The CIS–CEA–UPA evaluation framework, developed through prior experiments, provides a newsroom-centered benchmark for assessing AIGC-generated visuals. Instead of relying on general resolution or aesthetic scores, it emphasizes identity consistency, cultural fidelity, and publication appropriateness – qualities critical for journalism. Empirical testing shows that outputs meeting all three criteria are consistently preferred by editors for publication, whereas those failing any one dimension are more likely to be flagged for revision or rejection.

Moreover,this framework also serves as actionable feedback for iterative LoRA training, supporting the development of a sustainable Human-AI collaboration pipeline. Looking forward, this workflow holds potential for rapid adaptation across diverse news categories. By generalizing the annotation structure and training logic, it becomes feasible to construct a foundation model pre-trained on multi-domain journalistic corpora. Such a model could significantly reduce fine-tuning overhead and enable faster deployment for emerging news scenarios.Example can be also  given by a special coverage called 2025, A Beautiful Flower Gift of Xinhua News Agency that achieves wide dissemination and significant impact. The well trained LoRA and workflow raises the image production  efficiency by 25 \%,more importantly,as shown in Figure \ref{fig:DPL}it has received an excellent propagation effect for over 100 million reviewers nationwide.

\section{Conclusion and Discussion}
Moving past the direct comparison of two experiments, this study focuses on what the findings imply for AI-assisted special coverage making in journalism. The results indicate that today’s diffusion-based platforms, although strong in pure generative power, still need well-placed human guidance to satisfy newsroom standards of accuracy and accountability. When professional editors shape prompts, adjust LoRA weights, and review outputs in real time, the images show higher factual fidelity and stronger cultural alignment, a pattern that matches recent observations on responsible media technology \cite{lewis2025generative,trattner2022responsible}.

The CIS-CEA-UPA framework succeeds in turning broad editorial expectations into clear evaluation criteria. Images that meet identity coherence, contextual fidelity, and publication appropriateness gain higher credibility scores from experts and general audiences alike, a result that mirrors earlier work on interpretability and trust in AI \cite{doshi2017towards,ji2025interpretable}. Because these same metrics guide each round of LoRA fine-tuning, they form a practical loop for continual model improvement without undue cost.The pipeline is not a universal substitute for photojournalism.  It is contraindicated for events where visual veracity cannot be assured (e.g., legally sensitive reconstructions) or where composites could mislead.  In such cases, the SOP mandates abstention or explicit labeling.  The CIS-CEA-UPA checklist helps editors make these calls consistently.

Looking ahead, a promising step is to pre-train a lightweight foundation model on a wide, multi-domain news corpus. This approach could reduce the effort required to adapt the system to new beats such as climate reporting, local elections, and data-driven investigations, while maintaining the risk controls shown here. Closer links with newsroom content-management systems would also help editors spot compliance issues quickly and hand off tasks smoothly to automated tools. In addition, pairing visual evaluation with text-based fact-checking methods \cite{drolsbach2025characterizing,farooq2025deciphering} could create a more complete defense against misinformation. We hope these insights prompt researchers and practitioners to treat human judgment and algorithmic speed as complementary strengths instead of rival approaches.

Ultimately, this human-centered AI approach not only elevates the quality of machine-generated news visuals but also preserves the essential human values in journalism - truth-seeking, contextual understanding, and ethical responsibility - while harnessing the scalability of computational creativity.The pipeline is not a universal solution; in cases involving evidentiary use or sensitive individuals, abstention or disclosure is required. The CIS-CEA-UPA checklist helps editors make these calls consistently. The challenge ahead lies in standardizing these collaborative protocols across diverse media ecosystems without compromising editorial sovereignty or technological innovation.
The CIS-CEA-UPA checklist helps editors make these calls consistently. The challenge ahead lies in standardizing these collaborative protocols across diverse media ecosystems without compromising editorial sovereignty or technological innovation.
\section{Acknowledgments}
This research was made possible with support and collaboration of the Xinhua News Agency, originating from its practical workflow needs in the newsroom and aligning with operational standards. The refined workflow and key outcomes form an actionable case study, reflecting academic-industry synergy. The authors thank Haoran Jiang for his help with formatting, bug fixes, and advice to improve the manuscript.

\bibliography{aaai2026}

\clearpage
\end{document}